\documentclass{article}

\usepackage{microtype}
\usepackage{graphicx}
\usepackage{booktabs} % for professional tables

\usepackage{hyperref}

\usepackage{algorithmic}
\usepackage{amsmath}
\usepackage{xspace}
\usepackage{caption}
\usepackage{subcaption}
\usepackage{booktabs}
\usepackage{amssymb}
\usepackage{multirow}
\newcommand*{\argmax}{\ensuremath{\operatornamewithlimits{argmax}}\xspace}

\newcommand\BLEU{\textsc{Bleu}}

\newcommand\psp{p_\text{s2p}}
\newcommand\ppt{p_\text{p2t}}

\newcommand{\spDecoder}{\text{Decoder}_\text{s2p} }

\newcommand\Decoder{\text{Decoder}}

\DeclareMathOperator{\SCORE}{SCORE}

\usepackage[accepted]{icml2021}

\icmltitlerunning{Towards Reinforcement Learning for Pivot-based Neural Machine Translation with Non-autoregressive Transformer}

\begin{document}

\twocolumn[
\icmltitle{Towards Reinforcement Learning for Pivot-based Neural Machine Translation with Non-autoregressive Transformer}



\icmlsetsymbol{equal}{*}

\begin{icmlauthorlist}
\icmlauthor{Evgeniia Tokarchuk}{ebay,rwth}
\icmlauthor{Jan Rosendahl}{rwth}
\icmlauthor{Weiyue Wang}{rwth}
\icmlauthor{Pavel Petrushkov}{ebay}
\icmlauthor{Tomer Lancewicki}{ebay}
\icmlauthor{Shahram Khadivi}{ebay}
\icmlauthor{Hermann Ney}{rwth}
\end{icmlauthorlist}

\icmlaffiliation{ebay}{eBay Inc.} 
\icmlaffiliation{rwth}{Human Language Technology and Pattern Recognition Group,  RWTH Aachen University, Germany}

\icmlcorrespondingauthor{Evgeniia Tokarchuk}{e.tokarchuk@uva.nl}
%\icmlcorrespondingauthor{}{\texttt{\{ppetrushkov,tlancewicki,skhadivi\}@ebay.com}}
%\icmlcorrespondingauthor{}{\texttt{\{rosendahl,wwang,ney\}@cs.rwth-aachen.de}}

\icmlkeywords{Machine Translation, Reinforcement Learning}

\vskip 0.3in
]



\printAffiliationsAndNotice{}  % leave blank if no need to mention equal contribution

\begin{abstract}
Pivot-based neural machine translation (NMT) is commonly used in low-resource setups, especially for translation between non-English language pairs. It benefits from using high-resource source$\to$pivot and pivot$\to$target language pairs and an individual system is trained for both sub-tasks.
However, these models have no connection during training, and the source$\to$pivot model is not optimized to produce the best translation for the source$\to$target task.
In this work, we propose to train a pivot-based NMT system with the reinforcement learning (RL) approach, which has been investigated for various text generation tasks, including machine translation (MT). We utilize a non-autoregressive transformer and present an end-to-end pivot-based integrated model, enabling training on source$\to$target data.
\end{abstract}

\section{Introduction}

%Machine translation (MT) is an active 

Machine translation (MT) research is heavily focused on the investigation of language pairs that include English either as a source or as a target language.
Besides linguistic implications, translation between non-English language pairs frequently suffers from scarce direct parallel training data. 

Pivot-based MT~\cite{Woszczyna-1993-nbest,utiyama-isahara-2007-comparison} mitigates the problem of data scarcity by translating via a pivot language (usually English), i.e., along the language pairs source$\to$pivot (src$\to$piv) and pivot$\to$target (piv$\to$trg).
The pivot language is chosen in such a way that the src$\to$piv and piv$\to$trg models can be trained with more data than the direct source$\to$target (src$\to$trg) model.
Pivot-based models require a sequential, two-step decoding, meaning that an explicit intermediate pivot representation is obtained from the src$\to$piv model and then used as an input to the piv$\to$trg model.
%which might lead to the propagation of the decoding errors.
Note that the src$\to$piv model is optimized to produce the best possible src$\to$piv translation, not to generate a hypothesis which maximizes the performance of the full src$\to$trg model. 
%while the pivot-based translation aims to obtain the best src$\to$trg hypothesis.
Furthermore the pivot-based approach does not use the direct src$\to$trg training data, meaning that the system is not adapted to the desired task.

%Concatenation
Cascading of the src$\to$piv and piv$\to$trg models at training time is problematic since
%seems like straightforward solution. However, 
the gradient cannot be propagated through the $\argmax$ operation that generates the pivot hypothesis. 
Moreover, training two cascaded Transformer~\cite{Vaswani-2017-attention} models on src$\to$trg data is complicated by the fact that the decoder of the src$\to$piv model ($\Decoder_{s2p}$) requires the explicit pivot sequences as an input.
Since src-piv-trg three-way data is rare in practice this requires e.g. a search over the src$\to$piv model at each training step, which is computationally prohibitive.

In our work, we propose to (i) replace the autoregressive (AR) decoder of the src$\to$piv system with a non-autoregressive (NA) Transformer decoder~\cite{GB18} and (ii) apply a reinforcement learning (RL) approach to train the src$\to$piv model with direct feedback from the piv$\to$trg system on the src$\to$trg training data.
Although there is a performance gap between NA and AR approaches~\cite{GB18,ghazvininejad-etal-2019-mask}, the NA approach can generate a pivot hypothesis parallelizable without a computationally costly search procedure.
The NA approach allows us to generate pivot hypotheses on-the-fly at training time. To the best of our knowledge, we are the first ones who applied the RL approach for training pivot-based NMT jointly.
%the NA approach allows to use the sequence of unknown as an input to the decoder, and as a result, there is no need for explicit pivot sentences. 

\section{Related Work}
RL applications for neural machine translation (NMT) recently gathered interest in the research community~\cite{choshen2019weaknesses, wu-etal-2018-study,shen-etal-2016-minimum}. 
In these cases RL is used to directly optimize a metric such as \BLEU{}~\cite{papineni-etal-2002-bleu} or TER~\cite{Snover06astudy} on sentence-level in contrast to conventional maximum likelihood training. Policy optimization methods such as REINFORCE~\cite{williams1992reinforce} were successfully applied to various sequence generation tasks, including MT~\cite{ranzato2015sequence}. 

Previous works on pivot-based MT have already studied the possibility to jointly train the src$\to$piv and piv$\to$trg models~\cite{cheng-2017-jointpivot} or to transfer knowledge and parameters~\cite{kim-2019-pivot-transfer}.
Since translation between non-English languages in practice entails a problem of data scarcity, various methods have been proposed to leverage different data types. 
One of the most popular approaches is the use of multilingual machine translation~\cite{zhang-etal-2020-improving,aharoni-etal-2019-massively-mlt,johnson-etal-2017-googles-mlt} that is to transfer knowledge from one or more high-resource language pairs to the desired low-resource task.
Another popular technique considers monolingual target-side data~\cite{sennrich-etal-2016-improving, edunov-etal-2018-bt-at-scale} to generate synthetic end-to-end data. However, since this work focus on the application of the RL, we do not aim for comprehensive comparisons with multilingual NMT systems and various data augmentation strategies. We refer to~\cite{kim-2019-pivot-transfer} for in-depth comparison studies.

\section{Pivot-based NMT}
\label{sec:pivot-nmt}
Given the two independently trained sequence-to-sequence models
namely src$\to$piv ($\psp$) and piv$\to$trg ($\ppt$), the goal of pivot-based NMT is to find the target hypothesis $e_1^I$ given the source sentence $f_1^J$ which requires the intermediate pivot representation $z_1^K$.
Conventional pivot-based NMT does not involve training on the src$\to$trg data, and the models are connected in search via so-called two-step decoding:
\begin{align}
\label{eq:two-pass-decoding}
            \hat{z}_1^{\hat{K}} &=  \argmax_{K, z_1^K}{\prod_{k=1}^K \psp (z_k|z_1^{k-1}, f_1^J)} \\
            \hat{e}_1^{\hat{I}} &= \argmax_{I, e_1^I}{\prod_{i=1}^I \ppt (e_i|e_1^{i-1}, \hat{z}_1^{\hat{K}})}.
\end{align}
Although it is possible to decode multiple pivot hypotheses on the first step and apply re-ranking to obtain final target hypothesis~\cite{och-etal-1999-improved,cheng-2017-jointpivot}, usually only one pivot hypothesis is used as an input to the piv$\to$trg model.

\section{RL for Pivot-based NMT}
\label{sec:rl-pivot-nmt}
\begin{figure*}
\centering
    \subfloat[AR-based concatenated model.\label{fig:ar-concat}]{%
        \centering
         \includegraphics[width=0.7\linewidth]{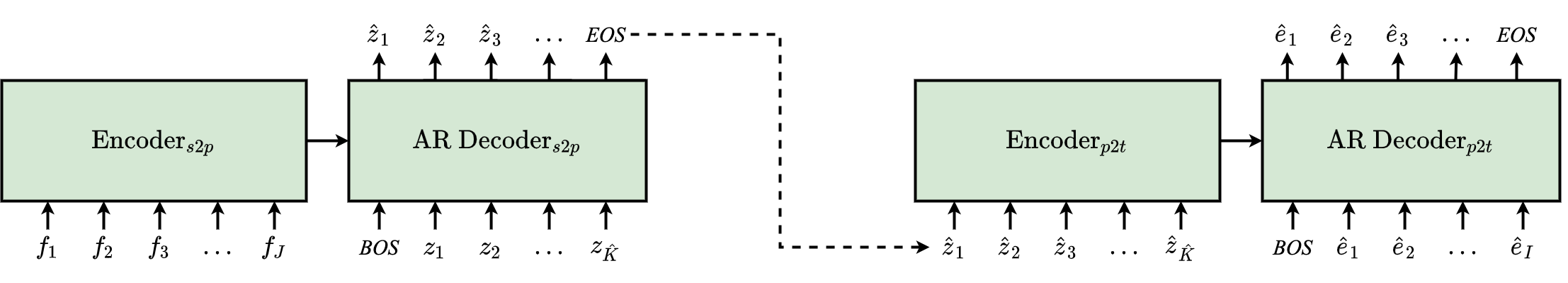}%
    }\par\vspace{0.3cm}
    \subfloat[NAT-based concatenated model.\label{fig:nat-concat}]{%
        \centering
         \includegraphics[width=0.7\linewidth]{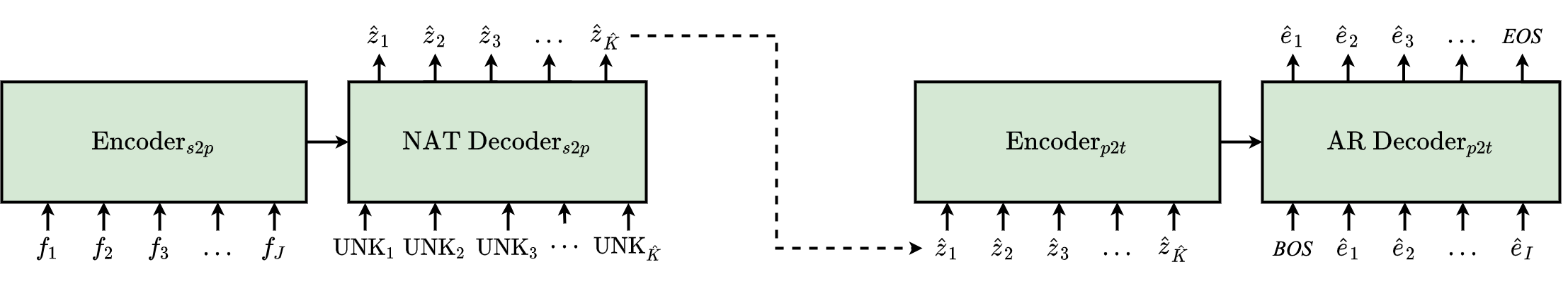}%
    }\par\vspace{0.3cm}
    \caption{Autoregressive (AR) and non-autoregressive transformer (NAT) architectures for pivot-based translation}
\end{figure*}

Starting with the setup described in Section~\ref{sec:pivot-nmt}, we aim to perform training on the src$\to$trg data to optimize the performance of the cascaded model.
However, in practice, src$\to$trg corpora do not provide a pivot reference which is needed to calculate the $\spDecoder$ as shown in Figure~\ref{fig:ar-concat}.
In such cases, synthetic pivot data can be obtained \lq offline\rq{} via (back-)translation \cite{sennrich-etal-2016-improving}, or \lq online\rq{} by performing a search for every update step of the training process.
For an offline generation, the pivot reference does not change, even if the parameters of the src$\to$piv model are updated and online updates are computationally too expensive. 
We replace the autoregressive decoder with a non-autoregressive one as shown in Figure~\ref{fig:nat-concat} which generates pivot hypothesis in parallel without the need for a search. That allows to speed up the sampling from the src$\to$piv model, as described in Appendix~\ref{app:sampling-speed}.
Specifically, we rely on the Conditional Masked Language Model (CMLM)~\cite{ghazvininejad-etal-2019-mask} as non-autoregressive model. Similar to~\cite{devlin-etal-2019-bert}, during training CMLM tries to predict the probabilities of masked tokens $Y_{mask}$ which depend on the source sequence and observed tokens $Y_{obs}$ of the target sequence $Y$ where $Y_{obs}=Y \setminus Y_{mask}$. The number of masked tokens is randomly chosen between one and sequence length. During decoding CMLM allows to perform multiple decoder iterations, which leads to the performance improvements~\cite{ghazvininejad-etal-2019-mask}. 
Since the CMLM model have seen fully-masked sequences during training, it allows us to use the sequence of unknown symbols as an input to the $\spDecoder$ and naturally solve the problem of missing pivot references in src$\to$trg data.

In pivot-based MT neither the src$\to$piv nor the piv$\to$trg model are optimized for the intended task. In this work we focus on improving the src$\to$piv model which is optimized for src$\to$trg performance on an extrinsic metric (typically BLEU).
Instead we propose to use reinforcement learning to train the src$\to$piv model to produce pivot hypotheses that the piv$\to$trg can translate well. 
For each training pair $(f_1 ^J, e_1 ^I)$ we aim to maximize the expected reward:
% In order to facilitate the training, we define the objective function on $\psp$ as an expected reward  as shown in Equation~\ref{eq:rl-obj}. 
\begin{gather}
  \label{eq:rl-obj}
  \mathcal{J}(\theta) = \mathbb{E}_{\psp(z_1^K|f_1^J)}[R (\ppt(\bullet | z_1^K), e_1^I) ]  
\end{gather}
where $R (\ppt(\bullet | z_1^K), e_1^I)$ denotes a reward function, which is based on the distribution of the piv$\to$ trg model $\ppt$ for the suggested pivot sample $z_1^K$. 

%The pivot sample is obtained by applying multionomial sampling on top of the $\psp$ model distribution. 

We consider two different types of reward. First, to use the negative cross-entropy (Neg. CE) between the target reference, interpreted as a one-hot distribution, and the distribution of the piv$\to$trg model $\ppt$:
\begin{equation}\label{eq:ce-reward}
        %R (\ppt(\bullet | z_1^K), e_1^I) = 
        - \sum_{i=1}^I \log ( \ppt(e_i|e_1^{i-1}, z_1^K) ).
\end{equation}
Second, we perform a search over the possible target hypotheses and utilize a sentence-level metric to compare the result $e_{\ppt} ( z_1 ^K) $ against the reference $e_1 ^I$:
\begin{gather}\label{eq:sl-reward}
        % R (\ppt(\bullet | z_1^K), e_1^I) = 
        \SCORE(e_{\ppt} ( z_1 ^K), e_1^I)
\end{gather}
where $\SCORE$ can be any sentence-level metric. Although in MT the common evaluation metric is corpus-level \BLEU{}~\cite{papineni-etal-2002-bleu}, in this work we consider sentence-level \BLEU{} and sentence-level chrF~\cite{popovic-2015-chrf}, which is another well-known MT evaluation metric. 
Since the objective in Equation~\ref{eq:rl-obj} is infeasible to compute in practice, we apply the REINFORCE~\cite{williams1992reinforce} algorithm to estimate the gradients. We define our action space as a target language vocabulary, and selecting the action is equivalent to sampling from the $\psp$ output distribution. We use multinomial sampling to generate pivot hypothesis during training, which can be done in parallel.
The training algorithm is formalized in Algorithm~\ref{alg:pivot-RL} and we use the length model provided by the CMLM for the src$\to$piv system.

\begin{algorithm}
\begin{algorithmic}
%\DontPrintSemicolon
\STATE \textbf{Input}: $\psp,\ \ppt, \text{length model } \ell _{\text{s2p}}, \text{corpus } \mathcal{C}$
 \FOR{$(f_1^J, e_1^I) \in \mathcal{C}$}
%   Predict pivot sentence length: $\hat{K}:=\argmax_{K}\psp(K|f_1^J)$\;
  \STATE Predict pivot sentence length: $\hat{K}:= \ell_{s2p}(f_1^J)$ 
  \STATE Sample pivot sequence: $\hat{z}_1^{\hat{K}} \sim \psp(z_1^K|f_1^J)$ 
  \STATE Generate target hypothesis: $\hat{e}_1^{\hat{I}} := \argmax \ppt ( \bullet|\hat{z}_1^{\hat{K}})$ 
  \STATE Score pivot sequence: $r := R (\ppt(\hat{e}_1^{\hat{I}} | \hat{z}_1^{\hat{K}}), e_1^I)$ 
  %R (\ppt(\hat{z}_1^{\hat{K}}, e_1^I) )$\;
  \STATE Perform gradient update:
  \STATE $\theta_{s2p}^{t+1}=\theta_{s2p}^{t} + \eta \nabla_{\theta_{s2p}} \log \psp(\hat{z}_1^{\hat{K}}|f_1^J) r$ 
  \ENDFOR
\caption{Pivot-based NMT with RL training}
\label{alg:pivot-RL}
\end{algorithmic}
\end{algorithm}

\section{Experiments}
\subsection{Data}
We validate the approach described in Section~\ref{sec:rl-pivot-nmt} on two translation datasets, which do not contain English as a source or target language: French$\to$German (Fr$\to$De) with 2.3M sentences and German$\to$Czech (De$\to$Cs) with 230K sentences of direct data from the WMT 2019 translation task\footnote{http://www.statmt.org/wmt19/}. We apply a \textit{byte-pair encoding} (BPE)~\cite{sennrich-etal-2016-neural} model with 32000 merge operations for each dataset.
Detailed data description and preprocessing steps can be found in Appendix~\ref{app:data}

\subsection{Model Training}
We use the fairseq\footnote{\url{github.com/pytorch/fairseq}} framework for all our experiments and compare the results to three different baselines:
\begin{itemize}
\itemsep0em
    \item \textbf{direct baseline}: a 'base`  Transformer~\cite{Vaswani-2017-attention} trained only on the on direct src$\to$trg data
    \item \textbf{pivot baseline}: Cascading a $\psp$ and $\ppt$ are 'base` Transformer model, each trained on src$\to$piv and piv$\to$trg data respectively.
    \item \textbf{NAT pivot baseline}: $\psp$ is a CMLM model (non-autoregressive) trained on the src$\to$piv data and $\ppt$ is a Transformer model (autoregressive) trained on piv$\to$trg data. Both models are individually trained and are only cascaded in src$\to$trg decoding. 
\end{itemize}

For all models based on the CMLM we set the effective batch size to be 65K tokens. The learning rate varies between $10^{-6}$ and $10^{-5}$. We utilize the Adam optimizer~\cite{kingma2015-adam} with $\beta=\{(0.9,0.98)\}$. 
Dropout is set to 0.1 for both language pairs.
To perform the validation steps with the CMLM, we apply five decoding iterations for the CMLM decoder.
If sentence-level BLEU scores are needed we use the SacreBLEU~\cite{post-2018-call} library.

\subsection{Results}

\begin{table*}[ht]
\centering
\resizebox{0.66\linewidth}{!}{  
\begin{tabular}{clcccccc}
\toprule
 & &  & \multicolumn{2}{c}{French$\rightarrow$German} && \multicolumn{2}{c}{German$\rightarrow$Czech}\\ \cline{4-5} \cline{7-8}
 \multicolumn{2}{l}{Method} & RL reward    &\multicolumn{2}{c}{$\BLEU^{[\%]}$} && \multicolumn{2}{c}{$\BLEU^{[\%]}$} \\
 &  &               & dev      & test      && dev      & test      \\  \midrule
\parbox[t]{2mm}{\multirow{2}{*}{\rotatebox[origin=c]{90}{AR}}}& 
    direct baseline & - & 20.0 & 20.4 && 13.5 &  14.0 \\
 & pivot baseline & - &  19.5 & 20.7  && 18.8	& 18.1 \\ \midrule
\parbox[t]{2mm}{\multirow{4}{*}{\rotatebox[origin=c]{90}{NAT}}}
  & pivot baseline & - & 17.1  & 18.1  && 17.3  & 16.6  \\
 &  \qquad +RL & Neg. CE  &  18.7 & 19.7 & & 18.2  & 17.5 \\
 &  & BLEU  &  19.0 & 19.9 & & 18.3  & 17.6 \\
 &  & chrF  &  18.6  & 19.7 & & 18.1 & 17.3 \\
 \bottomrule
\end{tabular}
}
\caption{Results on the pivot-based NMT with different RL rewards. All pivot/cascaded models are pre-trained on the respective data. We use \texttt{newstest\{2011,2012\}} as dev and test respectively.}
\label{tab:results}
\end{table*}

We report the results for three different kinds of rewards (Neg. CE, \BLEU{} and chrF) in Table~\ref{tab:results} and compare them against the baselines.
We scale Neg. CE by the sentence length and additionally reduce it by factor 10 to avoid large loss numbers. 
\BLEU{} and chrF are calculated using SacreBLEU and remain unchanged during training. 
We report case-sensitive \BLEU{} scores, ranging from 0 to 100 while chrF ranges from 0 to 1. 
Our RL approach consistently outperforms the non-autoregressive pivot baseline by up to 1.6\% \BLEU{} on Fr$\to$De and 0.9\% \BLEU{} on De$\to$Cs. However, it is still under the performance of the autoregressive pivot baseline for both languages.

In reported experiments, we do not rely on Knowledge Distillation (KD)~\cite{kim-rush-2016-sequence}, even though it is typically used for training NA models. The detailed discussion on KD is presented in Appendix~\ref{app:KD}

% \begin{table*}[ht]
% \centering
% \begin{tabular}{cllccccc}
% \toprule
%  & &  & \multicolumn{2}{c}{French$\rightarrow$German} && \multicolumn{2}{c}{German$\rightarrow$Czech}\\ \cline{4-5} \cline{7-8}
%  \multicolumn{3}{l}{Method}   &\multicolumn{2}{c}{$\BLEU^{[\%]}$} && \multicolumn{2}{c}{$\BLEU^{[\%]}$} \\
%  &                   &                & dev      & test      && dev      & test      \\  \midrule
% \parbox[t]{2mm}{\multirow{2}{*}{\rotatebox[origin=c]{90}{AR}}}& 
%     \multicolumn{2}{l}{direct baseline} & 20.0 & 20.4 && 13.5 &  14.0 \\
%  & \multicolumn{2}{l}{pivot baseline} &  19.5 & 20.7  && 18.8	& 18.1 \\ \midrule
% \parbox[t]{2mm}{\multirow{4}{*}{\rotatebox[origin=c]{90}{NAT}}}
%   & \multicolumn{2}{l}{pivot baseline} & 17.1  & 18.1  && 17.3  & 16.6  \\
%  &  \qquad+ RL with reward: & Neg. Cross-Entropy  &  18.7 & 19.7 & & 18.2  & 17.5 \\
%  &  &BLEU  &  19.0 & 19.9 & & 18.3  & 17.6 \\
%  &  &chrF  &  18.6  & 19.7 & & 18.1 & 17.3 \\
%  \bottomrule
% \end{tabular}
% \caption{All pivot/cascaded models are pre-trained on the respective data. We use \texttt{newstest\{2011,2012\}} as dev and test respectively.}
% \label{tab:results}
% \end{table*}

Directly optimizing the \BLEU{} score yields the best performance among all reinforcement learning setups by a small margin and overall the performance of all rewards gives comparable results. 

\subsection{Pivot \BLEU{} vs. target \BLEU}
For analysis we construct a three-way test set from the data of the Tatoeba challenge~\cite{tiedemann-2020-tatoeba}, that provides both a pivot and target references for each source sentence, obtaining
%In total we obtain 
10K sentence triples for En$\to$De$\to$Fr, that are not seen in training.

\begin{figure}[h]
    \centering
    \includegraphics[width=0.8\linewidth]{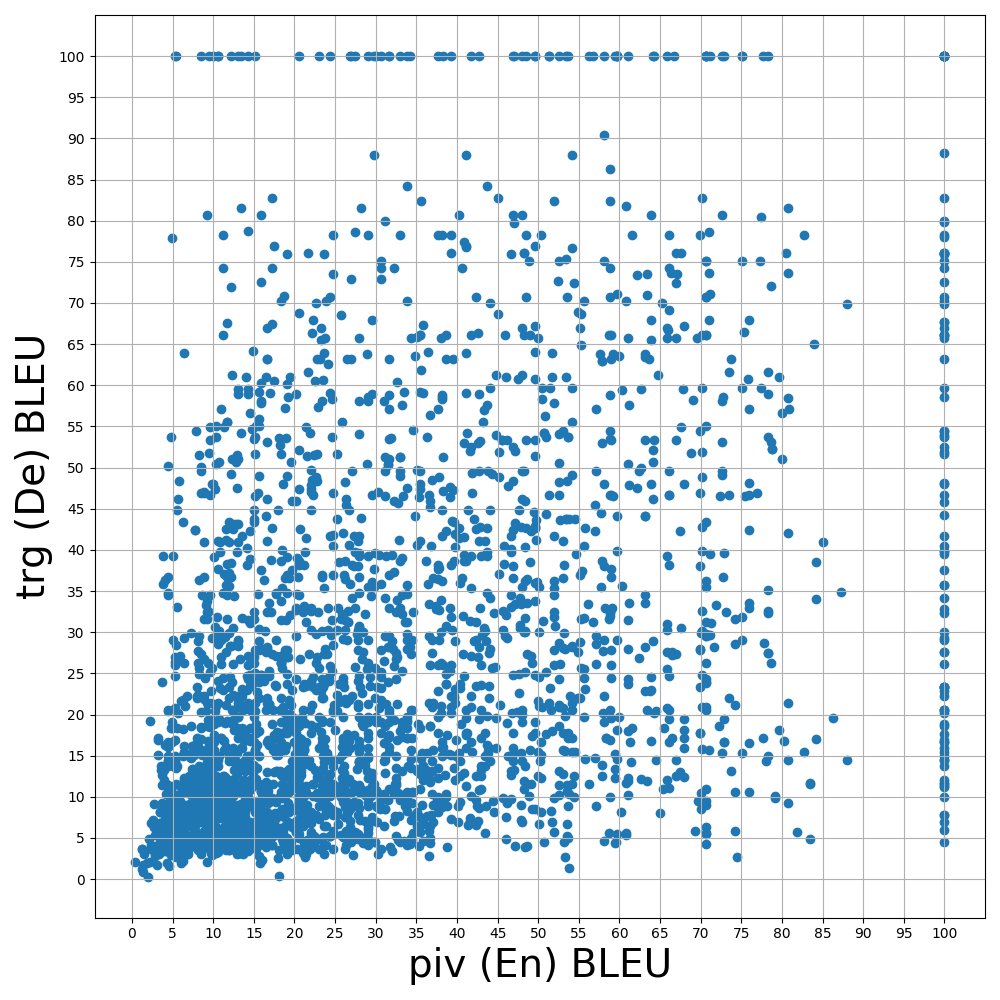}
    %\caption{Scatter plot of src$\to$piv sentence-level \BLEU{} versus src$\to$trg sentence-level \BLEU{}.}
    \caption{Sentence-level \BLEU{} score of the intermediate pivot hypothesis vs the sentence-level \BLEU{} of the target hypothesis. Both hypotheses are obtained from the same two step decoding for a pivot Fr$\to$En$\to$De system. }
    \label{fig:pivotBLeu-targetBLeu}
\end{figure}

On this test set we study the relationship between the pivot and target \BLEU{} score. 
Given the AR pivot baseline and NAT pivot baseline, we perform two-pass decoding obtaining a pivot and a target hypothesis as well as a sentence-level \BLEU{} score for each.
% We measure the sentence-level \BLEU{} score of the src$\to$piv hypothesis and piv$\to$trg hypothesis.
% Additionally, we use piv$\to$trg model and apply decoding on piv references. 
Figure~\ref{fig:pivotBLeu-targetBLeu} shows that src$\to$piv sentence-level \BLEU{} score does not correlate well with the src$\to$trg score. 
This indicates that simply improving src$\to$piv translation quality in terms of \BLEU{} will not necessarily result in the best src$\to$trg performance. Instead, optimizing src-$\to$piv directly for the src$\to$trg task may lead to the better model performance.

\section{Conclusion}
In this work, we propose a novel approach to pivot-based NMT. Applying reinforcement learning allows us to fine-tune a src$\to$piv model and obtain better performance of the cascaded src$\to$trg system.
%While this works in principle, this approach requires a weaker, non-autoregressive src$\to$piv system to be computationally feasible.
While this works in principle, our approach relies on a weaker, non-autoregressive src$\to$piv system and performance degradation of the NAT model could not be overcome and thus the conventional autoregressive pivot baseline still dominates in translation performance. However, we believe that the proposed method can be used in the pivot-based NMT and other cascaded models, like speech translation, where the hard decisions are required.
For future work, we suggest focusing on sampling and length modelling strategies for the non-autoregressive parts of the pipeline as well as investigating how autoregressives src$\to$piv models can be incorporated into the reinforcement pipeline.

% \section*{Acknowledgements}
\section*{Acknowledgments}
Authors affiliated with RWTH Aachen University have partially received funding from the European Research Council (ERC) (under the European Union's Horizon 2020 research and innovation programme, grant agreement No 694537, project ``SEQCLAS'') and eBay Inc. The work reflects only the authors' views, and none of the funding agencies is responsible for any use that may be made of the information it contains.

\bibliographystyle{icml2021}
\bibliography{icml2021,anthology}

\appendix
\section{Training Data}
\label{app:data}
Training data for French$\rightarrow$German includes Europarl corpus version 7~\cite{koehn2005europarl}, CommonCrawl\footnote{\url{https://commoncrawl.org/}} corpus and the newstest2008-2010. The original German$\rightarrow$Czech task was constrained to unsupervised translation, but we utilized the available parallel data to relax these constraints. The corpus consists of NewsCommentary version 14~\cite{TIEDEMANN12.463} and we extended it by including newssyscomb2009\footnote{\url{http://www.statmt.org/wmt09/system-combination-task.html}} and the concatenation of previous years test sets newstest2008-2010 from the news translation task. 
For both tasks we use newstest2011 as the development set and newstest2012 as the test sets. The data statistics, including pre-training data, are collected in Table~\ref{tab:data-stats}.

\begin{table}[ht]
    \centering
    \resizebox{0.75\linewidth}{!}{%
    \begin{tabular}{llcc}
    \toprule
    \multicolumn{2}{l}{}                  & \#Sent.   & \#Words (trg) \\ \hline
    direct data & Fr$\to$De & 2.3M & 53M \\ \hline 
    \multirow[t]{2}{*}{pre-train} & Fr$\to$En & 35M & 905M \\ 
    & En$\to$De & 9.7M & 221M \\ \hline\hline
    direct data & De$\to$Cs & 230K & 4.5M \\ \hline
    \multirow[t]{2}{*}{pre-train} & De$\to$En & 9.1M & 180M \\ 
    & En$\to$Cs & 49M & 486M \\ \bottomrule
    \end{tabular}%
    }
    \caption{Training data overview.}
    \label{tab:data-stats}
    \end{table}
For each parallel corpus, we apply a standard preprocessing procedure:
First, we tokenize each corpus using the Moses\footnote{\url{http://www.statmt.org/moses/}} tokenizer. Then a true-casing model is trained on all training data and applied to both training and test data. 
In the final step, we train a \textit{byte-pair encoding} (BPE)~\cite{sennrich-etal-2016-neural} model with 32000 merge operations.
The BPE models are trained BPE jointly on all available source, pivot and target data for the respective language triple.

\section{Knowledge Distillation}
\label{app:KD}
Non-autoregressive translation models are often trained using knowledge distillation (KD)~\cite{kim-rush-2016-sequence} to improve the model performance~\cite{ghazvininejad-etal-2019-mask,Zhou2020Understanding}.
%We apply knowledge distillation in the training of the src$\to$piv CMLM 
To strengthen the src$\to$piv CMLM we run an experiment in which we apply knowledge distillation in the training during the CMLM pre-training.
Although knowledge distillation does improve the performance of the src$\to$piv model, our experiments show that it does not lead to performance improvements on top of reinforcement learning in the end-to-end model. 
The resulting \BLEU{} score with and without knowledge distillation are depicted in Table~\ref{tab:res-KD}.

\begin{table}[ht]
\centering
\resizebox{0.75\linewidth}{!}{%
\begin{tabular}{lcc}
\toprule
Method  & \multicolumn{2}{c}{$\BLEU^{[\%]}$}  \\
        & Fr$\rightarrow$De & De$\rightarrow$Cs \\ \cline{1-3}
 NAT pivot baseline  & 17.1  &   17.3 \\
  \quad + knowledge dist. & 18.4  & 18.0    \\
  \quad + RL  &  19.0 & 18.3 \\
  \qquad + knowledge dist.  & 18.4  & 18.4  \\
 \bottomrule
\end{tabular}%
}
\caption{Results of the knowledge distillation on pivot systems with non-autoregressive src$\to$piv model. All results are reported on the development set.}
\label{tab:res-KD}
\end{table}

\section{Autoregressive vs. Non-autoregressive Sampling}
\label{app:sampling-speed}
To obtain samples for gradient estimation, we apply multinomial sampling on top of the model distribution as discussed in Section~\ref{sec:rl-pivot-nmt}.
In non-autoregressive model, the length is predicted beforehand and each token of the sampled sequence can be obtained in parallel with one decoder pass.
However, the sampling from the autoregressive model is not possible in parallel and has to be done left-to-right, which requires $\hat{K}_{max}$ passes of the decoder, where $\hat{K}_{max}$ is the maximum pivot sequence length. Thus, the training speed decreases significantly. To compare the sampling speed between non-autoregressive and autoregressive models, we measure the execution time for the sampling function using pytorch profiler\footnote{\url{https://pytorch.org/tutorials/recipes/recipes/profiler_recipe.html}}.
We do sampling on CMLM model and Transformer model using the validation set. On average, sampling with the non-autoregressive model takes 73.807 milliseconds while with the autoregressive model sampling takes 5.780 seconds, which makes non-autoregressive model more suitable in terms of training speed.
\end{document}